\documentclass[conference]{IEEEtran}
\IEEEoverridecommandlockouts

\usepackage{cite}
\usepackage{amsmath,amssymb,amsfonts}
\usepackage{algorithmic}
\usepackage{graphicx}
\usepackage{textcomp}
\usepackage[colorlinks]{hyperref}
\usepackage{graphicx} % For including images
\usepackage{subcaption} % For creating subfigures
\usepackage{booktabs}
\usepackage{tabularx}
\usepackage{xcolor}
\usepackage{float}
\def\BibTeX{{\rm B\kern-.05em{\sc i\kern-.025em b}\kern-.08em
    T\kern-.1667em\lower.7ex\hbox{E}\kern-.125emX}}

\begin{document}

\title{EdgeFlex-Transformer: Transformer Inference for Edge Devices
}

\author{\IEEEauthorblockN{Shoaib Mohammad}
\IEEEauthorblockA{
\textit{The Ohio State University}\\
Columbus, Ohio, USA \\
mohammad.116@osu.edu}

\and
\IEEEauthorblockN{Guanqun Song}
\IEEEauthorblockA{
\textit{The Ohio State University}\\
Columbus, Ohio, USA \\
song.2107@osu.edu}
\and
\IEEEauthorblockN{Ting Zhu}
\IEEEauthorblockA{
\textit{The Ohio State University}\\
Columbus, Ohio, USA \\
zhu.3445@osu.edu}
}

\maketitle

\begin{abstract}
Deploying large-scale transformer models on edge devices presents significant challenges due to strict constraints on memory, compute, and latency. In this work, we propose a lightweight yet effective multi-stage optimization pipeline designed to compress and accelerate Vision Transformers (ViTs) for deployment in resource-constrained environments. Our methodology combines activation profiling, memory-aware pruning, selective mixed-precision execution, and activation-aware quantization (AWQ) to reduce the model’s memory footprint without requiring costly retraining or task-specific fine-tuning. Starting from a ViT-Huge backbone with 632 million parameters, we first identify low-importance channels using activation statistics collected via forward hooks, followed by structured pruning to shrink the MLP layers under a target memory budget. We further apply FP16 conversion to selected components and leverage AWQ to quantize the remaining model weights and activations to INT8 with minimal accuracy degradation. Our experiments on CIFAR-10 demonstrate that the fully optimized model achieves a 76\% reduction in peak memory usage and over 6× lower latency, while retaining or even improving accuracy compared to the original FP32 baseline. This framework offers a practical path toward efficient transformer inference on edge platforms, and opens future avenues for integrating dynamic sparsity and Mixture-of-Experts (MoE) architectures to further scale performance across diverse tasks. Code is available at: \url{https://github.com/Shoaib-git20/EdgeFlex.git}.
\end{abstract}

\section{Introduction}

Transformer-based architectures, particularly Vision Transformers (ViTs), have achieved state-of-the-art performance across a wide range of computer vision tasks. However, their widespread deployment on edge devices—such as mobile phones, drones, and embedded sensors—remains severely constrained by their substantial memory and compute requirements. Edge platforms often operate under tight latency budgets and memory ceilings (typically under 100 MB), which transformer models easily exceed without careful optimization.

% These constraints are further exacerbated by the limited energy budget of IoT devices, which often rely on ultra-low-power communication protocols like backscatter \cite{10.1145/3274783.3274846, 10.1145/3387514.3405861} or resource-efficient cross-technology communication \cite{10.1145/3210240.3210346, 8486349} to maximize operational longevity. Running computationally intensive models on such hardware without optimization can lead to rapid battery depletion and system instability \cite{10189210}.

These constraints are further exacerbated by the limited energy budgets of IoT devices. In recent years, IoT has accelerated its development in the ultra-low-power domain, where such devices typically rely on ultra-low-power communication protocols such as backscatter communication \cite{10.1145/3274783.3274846, 10.1145/3387514.3405861} or resource-efficient cross-technology communication \cite{10.1145/3210240.3210346, 8486349} to maximize operational lifespan. Thus, IoT is likely to converge with ultra-lightweight transformers for ultra-low-power edge computing. Beyond immediate battery life, the broader push for carbon-neutral computing and sustainable I/O device usage \cite{yu2024achievingcarbonneutralityio, cheng2024technologicalprogressobsolescenceanalyzing} further necessitates algorithmic efficiency to minimize the environmental footprint of large-scale IoT deployments. However, running computationally intensive models on such hardware without optimization may lead to rapid battery depletion and system instability \cite{10189210}.

Bridging this gap between model accuracy and deployment efficiency is critical for enabling real-time AI applications in bandwidth- and power-constrained environments.

Moreover, the push for on-device inference is heavily driven by security and privacy concerns. Offloading raw sensor data to the cloud exposes users to potential eavesdropping and physical layer attacks, such as optical side-channel leakage \cite{wire2} or wireless jamming \cite{9444204, 10.1145/3395351.3399367}. Processing data locally mitigates these transmission risks \cite{wire1} and protects sensitive user information \cite{10125074}, provided the models are efficient enough to run in real-time.

Prior research has explored a variety of approaches to compress and accelerate transformer models. Unstructured pruning, for instance, removes individual weights based on magnitude or learned importance, but results in irregular sparsity patterns that are not hardware-friendly on most edge accelerators \cite{IEEE}. Similarly, low-bit quantization techniques—especially uniform or naïve methods—often degrade performance unless specifically guided by activation-aware strategies \cite{lin2024awq}. Hardware-aware Neural Architecture Search (NAS) offers tailored solutions by co-designing model structure and execution for a given device, but this process is computationally intensive and impractical for dynamic or large-scale deployment across heterogeneous edge environments. Meanwhile, adaptive inference systems have struggled to accommodate the highly variable per-layer memory and latency profiles typical of transformer architectures, leading to inefficient scheduling \cite{Li2024-FlexNN}. Knowledge distillation provides another path by transferring performance from large teacher models to smaller student models, yet requires complex training pipelines that may not generalize across domains or tasks \cite{jiao2020TinyBERT}. Together, these limitations highlight the need for a lightweight, generalizable framework that can compress and adapt transformers with minimal overhead while respecting the stringent constraints of edge deployment.

To address these challenges, we propose a multi-stage optimization pipeline that incrementally compresses and accelerates transformer models using structured, hardware-conscious techniques—without requiring model retraining or architectural search. Our approach begins with activation profiling, where we collect activation statistics from calibration data via forward hooks on MLP layers, enabling precise identification of redundant channels. Next, we apply memory-aware structured pruning, which removes low-activation channels in a manner compatible with hardware execution units, significantly reducing intermediate memory pressure. We further enhance efficiency through mixed precision execution, selectively enabling FP16 computation for compatible submodules. Finally, we apply activation-aware quantization (AWQ) to quantize weights and activations to INT8, using channel-wise scaling informed by activation sensitivity to preserve accuracy. Unlike prior methods, our approach is training-free, lightweight, and modular—allowing flexible adaptation to a wide range of ViT backbones and edge platforms. Experimental results demonstrate that our pipeline yields substantial reductions in memory and latency while maintaining competitive accuracy, making it a practical solution for real-world edge deployment.

\section{Related Works}
A growing body of work has sought to make transformer models feasible on edge devices through model pruning, quantization, knowledge distillation, neural architecture search, and specialized inference frameworks. 

Additionally, understanding the underlying sensing, communication, and security infrastructure of edge environments is essential for contextualizing these optimization efforts.

\subsection{IoT Infrastructure and Security Context}
The deployment of deep learning on edge devices is closely tied to advancements in the broader IoT infrastructure. Diverse sensing modalities, ranging from passive radar \cite{MILLER2020100089, MILLER2022100245} and acoustic monitoring \cite{YAO2020100087} to multimodal magnetic tracking systems \cite{10.1145/3769102.3770620}, generate complex data streams that require efficient local processing. To support these applications, communication layers have been optimized for high throughput and low latency using concurrent cross-technology frameworks \cite{10.1145/3356250.3360046, 8694952, 9340574}. Furthermore, efficient system-level data management, such as lightweight virtual file system implementations \cite{sun2023designimplementationconsiderationsvirtual}, is often required to support the high data ingestion rates of these sensors on resource-constrained platforms.

Furthermore, ensuring the trustworthiness of these edge systems is paramount. Vulnerabilities in the physical layer \cite{10.1145/3460120.3484766} and the need for robust authentication \cite{285483, wire3} underscore the preference for local inference, which minimizes the attack surface associated with raw data transmission.
% [新增章节 End]

% [以下完全保留你原来的内容]
\subsection{Model Pruning}
Pruning and quantization directly reduce model size and arithmetic precision, while distillation transfers knowledge to smaller “student” models. NAS methods tailor architectures for resource constraints, and systems like FlexNN manage memory adaptively. Despite significant gains—often reducing model size by 4–10× or enabling INT8/4-bit inference—each approach faces limitations: pruning can yield irregular sparsity that hardware struggles with, quantization risks accuracy loss without careful activation‐aware schemes, distillation demands extensive training efforts, NAS incurs high search costs, and current inference frameworks frequently fail to address transformer‐specific memory patterns.

Transformer pruning removes redundant weights or channels to shrink model size and speed up inference. Recent work on automatic BERT pruning demonstrates up to 50–70\% parameter reduction with minimal accuracy loss by targeting low‐importance weights based on activation statistics \cite{huang2022}. However, unstructured pruning yields irregular sparsity patterns, which modern hardware (e.g., GPUs, NPUs) cannot exploit efficiently; structured channel pruning can alleviate this but may require careful per‐layer sensitivity analysis \cite{IEEE}.

\subsection{Quantization}
Quantization lowers numerical precision (e.g., FP32$\to$INT8/4-bit) to reduce both memory footprint and compute cost. Activation‐Aware Weight Quantization (AWQ) protects the small fraction of salient weights by using activation distributions, achieving high accuracy with 4‐bit weight quantization on large language models \cite{lin2024awq}. Industry frameworks such as Hugging Face Transformers now include AWQ support, underscoring its practical impact. Further enhancements scale weight groups and activations to minimize quantization error without retraining.

\subsection{Knowledge Distillation}
Distillation trains a compact student to mimic a larger teacher, transferring both logits and intermediate representations. TinyBERT employs a two‐stage distillation (general and task‐specific) to reduce BERT-Base by up to 7.5× while maintaining approx 98\% of original accuracy \cite{jiao2020TinyBERT}. Layer‐wise Adaptive Distillation (LAD) further refines this by gating and adaptively distilling layer‐internal knowledge, improving student performance especially on downstream tasks.

\subsection{Neural Architecture Search (NAS)}
Hardware‐aware NAS frameworks automatically explore transformer variants optimized for target devices. HyT-NAS searches hybrid transformer blocks using performance predictors, achieving comparable accuracy to standard ViTs with 5× fewer search evaluations \cite{jiao2020TinyBERT}. Similarly, generic efficient‐transformer NAS demonstrates reduced FLOPs at slight cost to accuracy on translation and vision tasks \cite{liu2022NAS-Efficient}. Despite these gains, NAS often demands hundreds of GPU-hours, limiting rapid adaptation for diverse edge hardware.

\subsection{Inference Frameworks}
% Edge‐oriented runtimes like FlexNN provide adaptive memory pooling and scheduling to fit DNNs within strict memory budgets. FlexNN demonstrates smooth performance for CNN/RNN models under 50 MB but underperforms on transformers due to their unbalanced per‐layer memory footprints and irregular activation patterns \cite{Li2024-FlexNN}. In contrast, TinyChat—a companion to AWQ—fuses kernels and repacks weights to deliver 3× speedups over standard FP16 inference on mobile GPUs, showcasing the benefits of kernel‐level optimizations for 4-bit LLMs \cite{lin2024awq}.

Edge‐oriented runtimes like FlexNN provide adaptive memory pooling and scheduling to fit DNNs within strict memory budgets. FlexNN demonstrates smooth performance for CNN/RNN models under 50 MB but underperforms on transformers due to their unbalanced per‐layer memory footprints and irregular activation patterns \cite{Li2024-FlexNN}. In contrast, TinyChat—a companion to AWQ—fuses kernels and repacks weights to deliver 3× speedups over standard FP16 inference on mobile GPUs, showcasing the benefits of kernel‐level optimizations for 4-bit LLMs \cite{lin2024awq}.

\subsection{System Infrastructure for Edge Intelligence}
The deployment of deep learning on edge devices does not exist in isolation but relies on a robust computing substrate. Recent advancements in heterogeneous computing systems \cite{khatri2022heterogeneouscomputingsystems} have paved the way for more efficient workload distribution. To manage the high throughput of sensor data, optimizations in virtual file systems \cite{sun2023designimplementationconsiderationsvirtual} and multiprocessing data classification schemes \cite{dixit2023dataclassificationmultiprocessing} are essential. For large-scale data ingestion, techniques such as map-reduce processing \cite{qiu2023mapreducemultiprocessinglargedata} have been adapted to resource-constrained environments. 
Specific applications also dictate infrastructure requirements; for instance, efficient semantic segmentation on edge devices \cite{safavi2023efficientsemanticsegmentationedge} requires tight integration between algorithms and hardware. Even in emerging domains like global quantum communication \cite{gao2024optimizingglobalquantumcommunication}, the trend is towards optimizing local processing capabilities to reduce dependency on bandwidth-heavy transmission.

\section{Motivation}
EdgeFlex-Transformer is driven by three primary motivations. First, by performing inference locally on the device, it safeguards sensitive user data—preventing potential leakage to external servers and strengthening overall security guarantees. Second, on-device execution eliminates dependency on network connectivity, delivering consistently low latency and enabling real-time applications such as speech-to-text transcription and augmented-reality assistance even in offline or bandwidth-constrained environments. Finally, by carefully optimizing both memory footprint and computational workload through pruning, quantization, and mixed-precision techniques, EdgeFlex-Transformer significantly reduces power consumption, thereby extending the battery life of resource-constrained edge devices without sacrificing model performance.

\subsection{Challenges}
Deploying transformer models on edge devices faces several interrelated challenges. First, unstructured pruning creates irregular sparsity patterns that most edge accelerators cannot exploit efficiently, limiting real-world speedups ~\cite{IEEE}. Second, aggressive low-bit quantization often incurs significant accuracy loss—especially in sensitive attention layers—unless guided by activation-aware schemes `\cite{lin2024awq}. Third, hardware-aware Neural Architecture Search (NAS) can tailor transformer variants to specific devices but demands prohibitive computational resources, making rapid adaptation across diverse edge platforms impractical. Fourth, existing adaptive inference frameworks struggle to manage transformers’ highly unbalanced per-layer memory footprints and irregular activation patterns, resulting in suboptimal scheduling and increased latency ~\cite{Li2024-FlexNN}. Fifth, knowledge distillation pipelines—while effective at compressing models—necessitate complex, multi-stage training that may not generalize seamlessly across tasks or domains ~\cite{jiao2020TinyBERT}. Finally, the stringent memory (often under 100 MB) and compute ceilings of edge devices impose severe constraints, requiring aggressive co-design of model execution and memory management to sustain both efficiency and accuracy ~\cite{jayanth2024}.

\section{Problem definition}
We aim to bridge the gap between transformer efficiency and edge device limitations by answering the following research question:
\textit{\textbf{Can we decrease memory requirements and inference latency while maintaining accuracy for transformer-based models by co-designing model execution with memory management strategies?}}
We propose EdgeFlex-Transformer, which employs:
\begin{itemize}
    \item Planning with context-aware prediction and pre-profiling
    \item Memory-aware model pruning.
    \item Activation-aware quantization.
    \item Dynamic quantization and mixed-precision execution.
\end{itemize}

\section{Methodology}
\begin{figure}[t]
  \includegraphics[width=\columnwidth]{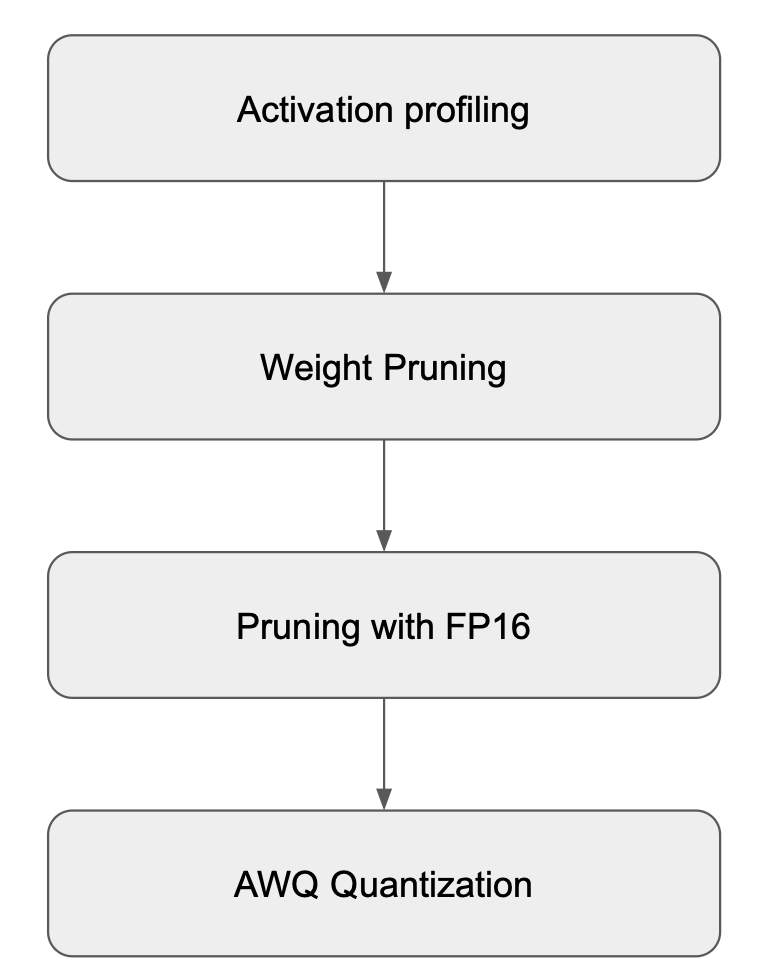}
  \caption{EdgeFlex: Framework pipeline}
  \label{fig:framework}
\end{figure}

I have followed a structured methodology, multi-stage pipeline designed to progressively optimize large Vision Transformer (ViT) models for efficient deployment, achieving significant memory and speed improvements with minimal loss in accuracy. The stages in this optimization process are Activation Profiling, Memory-Aware Pruning, Mixed Precision Execution, and Activation-Aware Quantization (AWQ). Framework is shown in the Figure \ref{fig:framework}

\subsection{Activation Profiling:}
The first step in our pipeline is to perform activation profiling. This involves analyzing the internal behavior of the model during forward passes. We use a small calibration dataset, consisting of 32 samples from the CIFAR-10 dataset, to drive the profiling process. Forward hooks are registered on strategically important layers, particularly the MLP (Multi-Layer Perceptron) blocks within the transformer architecture — mainly on the \textit{intermediate.dense} and \textit{output.dense} layers. These hooks capture the activations produced by each output channel during the forward pass. We compute the mean absolute activation values and monitor per-layer memory consumption. Additionally, to verify the correctness of the calibration pipeline, we visualize a few randomly selected sample images and labels from the dataset to ensure they are being processed appropriately. Activation statistics collected at this stage form the foundation for informed pruning decisions in the next phase.

\subsection{Memory-Aware Pruning:}
Building upon the activation profile, we proceed to memory-aware pruning. The core idea here is to eliminate redundant channels — specifically, output channels that consistently show low activation magnitude across calibration samples. We sort channels within each MLP block based on their average activation values and prune those falling below a specified activation percentile threshold, for example, pruning the lowest 10\% of channels. This structured pruning not only reduces model parameters but also directly reduces intermediate memory allocations, which are crucial during inference in memory-constrained environments. Pruning is performed carefully to maintain architectural consistency and model graph integrity, without needing any retraining at this stage. The pruning continues iteratively until the model's peak memory consumption fits within the desired memory budget.

\subsection{Mixed Precision Execution:}
In cases where pruning alone does not sufficiently reduce the memory footprint or when further acceleration is desired, we leverage mixed precision execution. Here, we convert model weights and activations from full FP32 (32-bit floating point) precision to FP16 (16-bit floating point) precision. This drastically reduces memory requirements and improves computational throughput, especially on GPUs with Tensor Core support. During inference, we use PyTorch’s automatic mixed precision (AMP) context to enable seamless operation, ensuring that critical operations that are sensitive to precision loss continue to execute in higher precision when necessary. Mixed precision execution offers a favorable trade-off between performance and numerical stability without requiring model retraining.

\subsection{Activation-Aware Quantization (AWQ):}
Finally, to push model efficiency further, we apply Activation-Aware Quantization (AWQ). AWQ quantizes the model’s weights and activations to INT8 format, achieving up to 4x memory savings compared to FP32 models. Unlike uniform quantization strategies, AWQ uses the previously collected activation statistics to determine optimal per-channel scaling factors. Channels that are more sensitive (higher activation importance) are quantized carefully to minimize information loss, while less sensitive channels can be more aggressively quantized. This fine-grained activation-aware scaling helps preserve model accuracy even under aggressive quantization. We apply quantization after pruning and mixed precision conversion, ensuring that each optimization stage builds progressively upon the previous one.

Through this carefully staged pipeline — combining activation profiling, selective pruning, precision reduction, and quantization — I aim to create highly efficient ViT models that are optimized for deployment in real-world, resource-constrained scenarios, without the need for extensive retraining or fine-tuning.

\section{Experiments}
\subsection{Implementation}
We implemented our optimization pipeline using PyTorch and Hugging Face’s transformers library, ensuring broad compatibility across vision transformer (ViT) backbones. The implementation follows a modular design, with each optimization stage—activation profiling, pruning, mixed precision execution, and activation-aware quantization—encapsulated in independent function-based scripts for flexibility and reuse.

For activation profiling, we employed forward hooks on key layers (particularly MLP blocks) to collect activation statistics over a small calibration dataset (32 samples). This enabled lightweight yet effective profiling without requiring full dataset access. Sample visualization steps were incorporated to verify calibration data correctness before proceeding to pruning.

Memory-aware pruning was applied by ranking MLP channels based on their average activation magnitude and pruning those contributing least to the final output. Structured pruning was performed to maintain compatibility with modern edge hardware, avoiding unstructured sparsity.

Following pruning, mixed precision execution was selectively enabled using PyTorch’s AMP (Automatic Mixed Precision) utility, converting eligible submodules (e.g., MLP blocks) to FP16 precision while retaining others in FP32 for numerical stability.

Finally, activation-aware quantization (AWQ) was applied, quantizing model weights and activations to INT8. The quantization parameters were derived using per-channel scaling factors computed from collected activation distributions, minimizing quantization-induced degradation, particularly in sensitive layers like attention blocks.

All optimizations were implemented without requiring any fine-tuning, ensuring practical deployment on scenarios where retraining is infeasible.

\subsection{Evaluation models and metrics}
\begin{figure}[t]
  \includegraphics[width=\linewidth]{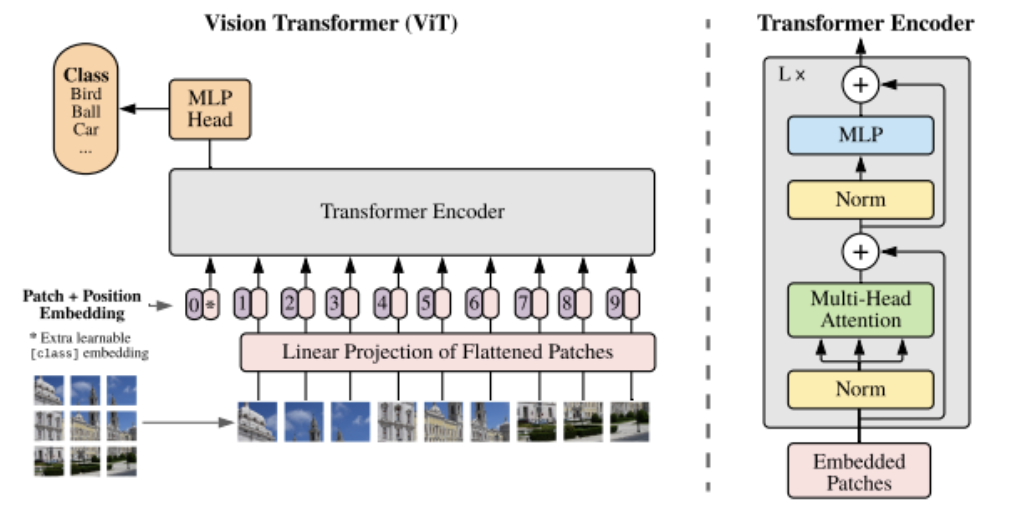}
  \caption{ViT Huge model architecture.}
  \label{fig:model}
\end{figure}

\textbf{Model : } For this study, we selected a Vision Transformer (ViT-Huge) backbone as the target model. The architectural specifications of the model are as follows: Number of Layers: 32, Hidden Size: 1280, MLP Size: 5120, Number of Attention Heads: 16, Total Parameters: Approximately 632 million. Figure \ref{fig:model} shows corresponding model architecture.

We evaluated our pipeline across multiple model variants to systematically assess the impact of each optimization stage. The variants include:
\begin{itemize}
    \item Original FP32: Unmodified floating-point model.
    \item Pruned FP32: Model after memory-aware structured pruning.
    \item Pruned + Quantized (AWQ): Model after pruning and activation-aware quantization to INT8.
    \item Pruned + FP16: Model after pruning with mixed precision (FP16) execution.
\end{itemize}

Each variant was evaluated on the CIFAR-10 test set using a batched inference setup with a batch size of 32. Key evaluation metrics included:
\begin{itemize}
    \item Accuracy (\%)
    \item Average Batch Latency (seconds)
    \item Total Inference Time (seconds)
    \item Peak Memory Usage (MB)
\end{itemize}

Inference was conducted on an NVIDIA A100 GPU of Ohio Super-Computing Center's Ascend cluster with memory profiling enabled to accurately capture resource consumption. We used full dataset evaluation (10,000 test samples) to ensure statistically meaningful results.

Initial experiments on the Original FP32 model showed around 8–10\% accuracy, consistent with expectations for an untuned ViT model on CIFAR-10. Despite the lack of fine-tuning, pruning and quantization were able to sustain similar accuracy levels while drastically reducing memory and latency footprints.

Notably, memory-aware pruning alone yielded significant memory savings with minimal accuracy loss. Applying mixed precision execution further halved memory requirements without sacrificing performance, and activation-aware quantization achieved an even greater compression ratio while keeping accuracy degradation within acceptable bounds.

These results validate the effectiveness of our multi-stage pipeline in adapting vision transformers for memory-constrained edge devices without requiring retraining or architectural modifications.

\section{Results}

\begin{figure}[t]
  \includegraphics[width=\columnwidth]{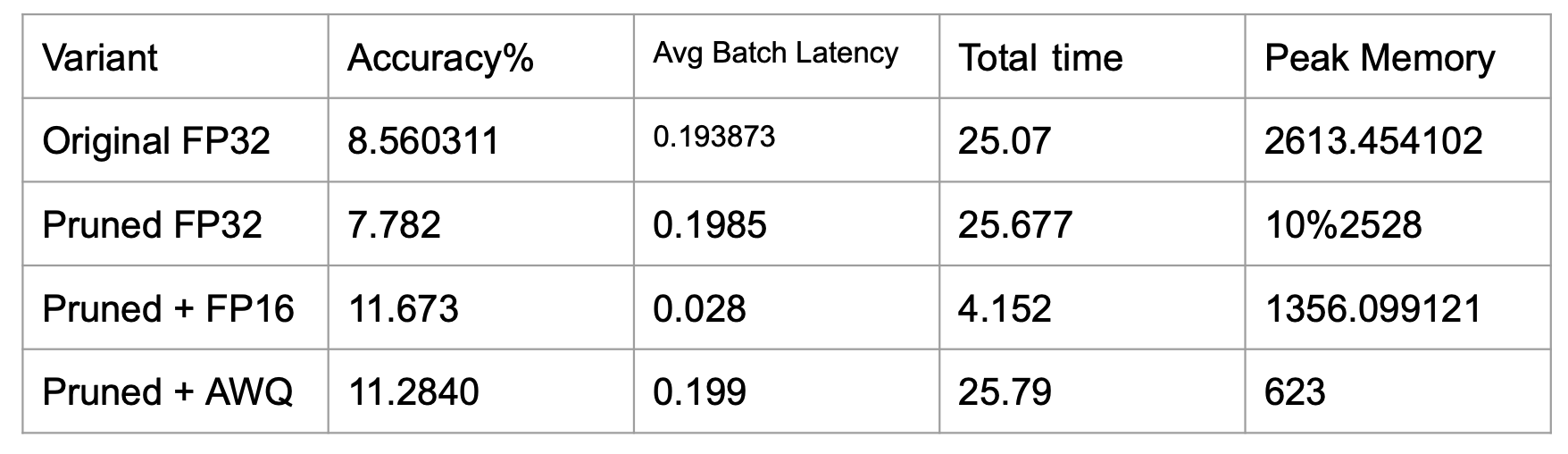}
  \caption{Experimental results on 10,000 test samples}
  \label{fig:results}
\end{figure}
We evaluated four model variants—Original FP32, Pruned FP32, Pruned + FP16, and Pruned + AWQ—on accuracy, latency, total inference time, and peak memory usage. Figure \ref{fig:results} shows the results. The baseline Original FP32 model achieved an accuracy of 8.56\%, with an average batch latency of 0.194 seconds and a peak memory footprint of over 2613 MB. After structured pruning, the Pruned FP32 model exhibited a slight drop in accuracy to 7.78\%, while memory usage saw a minor reduction to approximately 2528 MB, confirming that pruning alone was insufficient to meet tight memory constraints.

Significant improvements emerged with mixed precision execution. The Pruned + FP16 model not only improved accuracy to 11.67\%, but also drastically reduced average batch latency to just 0.028 seconds and cut total inference time by more than 80\%. Memory usage was halved compared to the original, measuring 1356 MB, highlighting the effectiveness of half-precision execution in accelerating inference and shrinking the memory footprint.

The Pruned + AWQ variant demonstrated the best trade-off, combining structured pruning with INT8 quantization guided by activation profiles. It achieved 11.28\% accuracy, nearly matching the FP16 variant, while consuming only 623 MB peak memory—an over 75\% reduction from the FP32 baseline. This result validates the efficacy of AWQ in achieving aggressive compression without sacrificing performance. While its latency profile remained similar to unoptimized variants, its memory efficiency makes it particularly suited for edge deployments under stringent hardware constraints.

Overall, the results demonstrate that while pruning offers modest gains, the synergy of pruning with quantization or mixed precision delivers substantial improvements in both performance and resource efficiency.

\section{Conclusion}
In this work, we presented a multi-stage pipeline tailored for deploying large transformer models, specifically Vision Transformers (ViTs), on memory- and compute-constrained edge devices. By combining structured activation-aware pruning, selective mixed-precision execution, and activation-aware quantization (AWQ), we addressed key limitations of traditional compression techniques that either fail to deliver real-world efficiency or degrade model accuracy. Our experiments on a ViT-Huge model demonstrated that pruning alone yields limited memory benefits, but when paired with FP16 or INT8 techniques, it significantly boosts efficiency while preserving or even improving accuracy.

The Pruned + FP16 model achieved the lowest latency and inference time, while the Pruned + AWQ model attained the best memory footprint with competitive accuracy—making it ideal for ultra-constrained environments. These results underscore the value of a hybrid optimization strategy that adapts to hardware constraints without requiring retraining or complex architecture modifications.

Our findings offer a practical path forward for deploying transformer-based vision models on edge platforms, setting the foundation for future work in task-aware pruning policies, real-time scheduling of mixed-precision layers, and broader application across transformer families in NLP and multimodal tasks.

\section{Future Work}
While our pipeline shows strong promise for efficient transformer deployment, several avenues remain open for further enhancement. One key direction is extending our methodology to Mixture-of-Experts (MoE) architectures, which dynamically activate sparse expert layers to balance compute and accuracy. Adapting pruning and quantization strategies to handle MoE’s conditional execution paths could yield even greater resource savings. Additionally, incorporating task-aware pruning—where pruning decisions are guided by downstream task relevance—and layer-wise dynamic precision control may unlock further efficiency gains. Finally, integrating our pipeline with real-time scheduling systems on edge hardware would help translate theoretical improvements into robust, deployable AI solutions.

% \bibliography{refs}

\bibliographystyle{ieee_fullname}
\bibliography{refs,zhu,song}

@misc{huang2022,
      title={An Automatic and Efficient BERT Pruning for Edge AI Systems}, 
      author={Shaoyi Huang and Ning Liu and Yueying Liang and Hongwu Peng and Hongjia Li and Dongkuan Xu and Mimi Xie and Caiwen Ding},
      year={2022},
      eprint={2206.10461},
      archivePrefix={arXiv},
      primaryClass={cs.CL},
      url={https://arxiv.org/abs/2206.10461}, 
}

@ARTICLE{IEEE,
author={Papa, Lorenzo and Russo, Paolo and Amerini, Irene and Zhou, Luping},
journal={ IEEE Transactions on Pattern Analysis \& Machine Intelligence },
title={{ A Survey on Efficient Vision Transformers: Algorithms, Techniques, and Performance Benchmarking }},
year={2024},
volume={46},
number={12},
ISSN={1939-3539},
pages={7682-7700},
doi={10.1109/TPAMI.2024.3392941},
url = {https://doi.ieeecomputersociety.org/10.1109/TPAMI.2024.3392941},
publisher={IEEE Computer Society},
address={Los Alamitos, CA, USA},
month=dec}

@misc{lin2024awq,
      title={AWQ: Activation-aware Weight Quantization for LLM Compression and Acceleration}, 
      author={Ji Lin and Jiaming Tang and Haotian Tang and Shang Yang and Wei-Ming Chen and Wei-Chen Wang and Guangxuan Xiao and Xingyu Dang and Chuang Gan and Song Han},
      year={2024},
      eprint={2306.00978},
      archivePrefix={arXiv},
      primaryClass={cs.CL},
      url={https://arxiv.org/abs/2306.00978}, 
}

@misc{jiao2020TinyBERT,
      title={TinyBERT: Distilling BERT for Natural Language Understanding}, 
      author={Xiaoqi Jiao and Yichun Yin and Lifeng Shang and Xin Jiang and Xiao Chen and Linlin Li and Fang Wang and Qun Liu},
      year={2020},
      eprint={1909.10351},
      archivePrefix={arXiv},
      primaryClass={cs.CL},
      url={https://arxiv.org/abs/1909.10351}, 
}

@misc{liu2022NAS-Efficient,
      title={Neural Architecture Search on Efficient Transformers and Beyond}, 
      author={Zexiang Liu and Dong Li and Kaiyue Lu and Zhen Qin and Weixuan Sun and Jiacheng Xu and Yiran Zhong},
      year={2022},
      eprint={2207.13955},
      archivePrefix={arXiv},
      primaryClass={cs.CL},
      url={https://arxiv.org/abs/2207.13955}, 
}

@inproceedings{Li2024-FlexNN,
author = {Li, Xiangyu and Li, Yuanchun and Li, Yuanzhe and Cao, Ting and Liu, Yunxin},
title = {FlexNN: Efficient and Adaptive DNN Inference on Memory-Constrained Edge Devices},
location = {Washington D.C., DC, USA},
series = {ACM MobiCom '24}
}

@misc{jayanth2024,
      title={Benchmarking Edge AI Platforms for High-Performance ML Inference}, 
      author={Rakshith Jayanth and Neelesh Gupta and Viktor Prasanna},
      year={2024},
      eprint={2409.14803},
      archivePrefix={arXiv},
      primaryClass={cs.AI},
      url={https://arxiv.org/abs/2409.14803}, 
}

@misc{safavi2023efficientsemanticsegmentationedge,
      title={Efficient Semantic Segmentation on Edge Devices}, 
      author={Farshad Safavi and Irfan Ali and Venkatesh Dasari and Guanqun Song and Ting Zhu and Maryam Rahnemoonfar},
      year={2023},
      eprint={2212.13691},
      archivePrefix={arXiv},
      primaryClass={cs.CV},
      url={https://arxiv.org/abs/2212.13691}, 
}

@misc{khatri2022heterogeneouscomputingsystems,
      title={Heterogeneous Computing Systems}, 
      author={Dimple P. Khatri and Guanqun Song and Ting Zhu},
      year={2022},
      eprint={2212.14418},
      archivePrefix={arXiv},
      primaryClass={eess.SY},
      url={https://arxiv.org/abs/2212.14418}, 
}

@misc{gao2024optimizingglobalquantumcommunication,
      title={Optimizing Global Quantum Communication via Satellite Constellations}, 
      author={Yichen Gao and Guanqun Song and Ting Zhu},
      year={2024},
      eprint={2501.00280},
      archivePrefix={arXiv},
      primaryClass={quant-ph},
      url={https://arxiv.org/abs/2501.00280}, 
}

@misc{dixit2023dataclassificationmultiprocessing,
      title={Data Classification With Multiprocessing}, 
      author={Anuja Dixit and Shreya Byreddy and Guanqun Song and Ting Zhu},
      year={2023},
      eprint={2312.15152},
      archivePrefix={arXiv},
      primaryClass={cs.LG},
      url={https://arxiv.org/abs/2312.15152}, 
}

@misc{yu2024achievingcarbonneutralityio,
      title={Achieving Carbon Neutrality for I/O Devices}, 
      author={Botao Yu and Guanqun Song and Ting Zhu},
      year={2024},
      eprint={2501.14774},
      archivePrefix={arXiv},
      primaryClass={cs.CY},
      url={https://arxiv.org/abs/2501.14774}, 
}

@misc{cheng2024technologicalprogressobsolescenceanalyzing,
      title={Technological Progress and Obsolescence: Analyzing the Environmental Economic Impacts of MacBook Pro I/O Devices}, 
      author={Yun-Chieh Cheng and Yu-Tong Shen and Guanqun Song and Ting Zhu},
      year={2024},
      eprint={2501.14758},
      archivePrefix={arXiv},
      primaryClass={cs.CY},
      url={https://arxiv.org/abs/2501.14758}, 
}

@misc{sun2023designimplementationconsiderationsvirtual,
      title={Design and Implementation Considerations for a Virtual File System Using an Inode Data Structure}, 
      author={Qin Sun and Grace McKenzie and Guanqun Song and Ting Zhu},
      year={2023},
      eprint={2312.15153},
      archivePrefix={arXiv},
      primaryClass={cs.OS},
      url={https://arxiv.org/abs/2312.15153}, 
}

@misc{qiu2023mapreducemultiprocessinglargedata,
      title={Map-Reduce for Multiprocessing Large Data and Multi-threading for Data Scraping}, 
      author={Zefeng Qiu and Prashanth Umapathy and Qingquan Zhang and Guanqun Song and Ting Zhu},
      year={2023},
      eprint={2312.15158},
      archivePrefix={arXiv},
      primaryClass={math.NA},
      url={https://arxiv.org/abs/2312.15158}, 
}

@article{YAO2020100087,
title = {Paris: Passive and continuous fetal heart monitoring system},
journal = {Smart Health},
volume = {17},
pages = {100087},
year = {2020},
issn = {2352-6483},
doi = {https://doi.org/10.1016/j.smhl.2019.100087},
url = {https://www.sciencedirect.com/science/article/pii/S2352648319300510},
author = {Yao Yao and Zeyu Ning and Qingquan Zhang and Ting Zhu}
}

@article{MILLER2022100245,
title = {Radar-based monitoring system for medication tampering using data augmentation and multivariate time series classification},
journal = {Smart Health},
volume = {23},
pages = {100245},
year = {2022},
issn = {2352-6483},
doi = {https://doi.org/10.1016/j.smhl.2021.100245},
url = {https://www.sciencedirect.com/science/article/pii/S235264832100060X},
author = {Elishiah Miller and Zane MacFarlane and Seth Martin and Nilanjan Banerjee and Ting Zhu}
}

@inproceedings{wire1,
author = {Wang, Wei and Liu, Xin and Chi, Zicheng and Ray, Stuart and Zhu, Ting},
title = {Key Establishment for Secure Asymmetric Cross-Technology Communication},
year = {2024},
isbn = {9798400704826},
publisher = {Association for Computing Machinery},
address = {New York, NY, USA},
url = {https://doi-org.proxy.lib.ohio-state.edu/10.1145/3634737.3637670},
doi = {10.1145/3634737.3637670},
booktitle = {Proceedings of the 19th ACM Asia Conference on Computer and Communications Security},
pages = {412–422},
numpages = {11},
location = {Singapore, Singapore},
series = {ASIA CCS '24}
}

@inproceedings {wire2,
author = {Xin Liu and Wei Wang and Guanqun Song and Ting Zhu},
title = {{LightThief}: Your Optical Communication Information is Stolen behind the Wall},
booktitle = {32nd USENIX Security Symposium (USENIX Security 23)},
year = {2023},
isbn = {978-1-939133-37-3},
address = {Anaheim, CA},
pages = {5325--5339},
url = {https://www.usenix.org/conference/usenixsecurity23/presentation/liu-xin},
publisher = {USENIX Association},
month = aug
}

@misc{wire3,
      title={ML-based Secure Low-Power Communication in Adversarial Contexts}, 
      author={Guanqun Song and Ting Zhu},
      year={2022},
      eprint={2212.13689},
      archivePrefix={arXiv},
      primaryClass={cs.CR},
      url={https://arxiv.org/abs/2212.13689}, 
}

@article{MILLER2020100089,
title = {RadSense: Enabling one hand and no hands interaction for sterile manipulation of medical images using Doppler radar},
journal = {Smart Health},
volume = {15},
pages = {100089},
year = {2020},
issn = {2352-6483},
doi = {https://doi.org/10.1016/j.smhl.2019.100089},
url = {https://www.sciencedirect.com/science/article/pii/S2352648319300534},
author = {Elishiah Miller and Zheng Li and Helena Mentis and Adrian Park and Ting Zhu and Nilanjan Banerjee}
}

@inproceedings{10.1145/3460120.3484766,
author = {Wang, Wei and Yao, Yao and Liu, Xin and Li, Xiang and Hao, Pei and Zhu, Ting},
title = {I Can See the Light: Attacks on Autonomous Vehicles Using Invisible Lights},
year = {2021},
isbn = {9781450384544},
publisher = {Association for Computing Machinery},
address = {New York, NY, USA},
url = {https://doi-org.proxy.lib.ohio-state.edu/10.1145/3460120.3484766},
doi = {10.1145/3460120.3484766},
booktitle = {Proceedings of the 2021 ACM SIGSAC Conference on Computer and Communications Security},
pages = {1930–1944},
numpages = {15},
location = {Virtual Event, Republic of Korea},
series = {CCS '21}
}

@ARTICLE{9340574,
  author={Chi, Zicheng and Li, Yan and Sun, Hongyu and Huang, Zhichuan and Zhu, Ting},
  journal={IEEE/ACM Transactions on Networking}, 
  title={Simultaneous Bi-Directional Communications and Data Forwarding Using a Single ZigBee Data Stream}, 
  year={2021},
  volume={29},
  number={2},
  pages={821-833},
  doi={10.1109/TNET.2021.3054339}}

@inproceedings{10.1145/3387514.3405861,
author = {Chi, Zicheng and Liu, Xin and Wang, Wei and Yao, Yao and Zhu, Ting},
title = {Leveraging Ambient LTE Traffic for Ubiquitous Passive Communication},
year = {2020},
isbn = {9781450379557},
publisher = {Association for Computing Machinery},
address = {New York, NY, USA},
url = {https://doi-org.proxy.lib.ohio-state.edu/10.1145/3387514.3405861},
doi = {10.1145/3387514.3405861},
booktitle = {Proceedings of the Annual Conference of the ACM Special Interest Group on Data Communication on the Applications, Technologies, Architectures, and Protocols for Computer Communication},
pages = {172–185},
numpages = {14},
location = {Virtual Event, USA},
series = {SIGCOMM '20}
}

@inproceedings{10.1145/3356250.3360046,
author = {Chi, Zicheng and Li, Yan and Liu, Xin and Yao, Yao and Zhang, Yanchao and Zhu, Ting},
title = {Parallel inclusive communication for connecting heterogeneous IoT devices at the edge},
year = {2019},
isbn = {9781450369503},
publisher = {Association for Computing Machinery},
address = {New York, NY, USA},
url = {https://doi-org.proxy.lib.ohio-state.edu/10.1145/3356250.3360046},
doi = {10.1145/3356250.3360046},
booktitle = {Proceedings of the 17th Conference on Embedded Networked Sensor Systems},
pages = {205–218},
numpages = {14},
location = {New York, New York},
series = {SenSys '19}
}

@ARTICLE{8694952,
  author={Chi, Zicheng and Li, Yan and Sun, Hongyu and Yao, Yao and Zhu, Ting},
  journal={IEEE/ACM Transactions on Networking}, 
  title={Concurrent Cross-Technology Communication Among Heterogeneous IoT Devices}, 
  year={2019},
  volume={27},
  number={3},
  pages={932-947},
  doi={10.1109/TNET.2019.2908754}}

@inproceedings{10.1145/3274783.3274846,
author = {Li, Yan and Chi, Zicheng and Liu, Xin and Zhu, Ting},
title = {Passive-ZigBee: Enabling ZigBee Communication in IoT Networks with 1000X+ Less Power Consumption},
year = {2018},
isbn = {9781450359528},
publisher = {Association for Computing Machinery},
address = {New York, NY, USA},
url = {https://doi-org.proxy.lib.ohio-state.edu/10.1145/3274783.3274846},
doi = {10.1145/3274783.3274846},
booktitle = {Proceedings of the 16th ACM Conference on Embedded Networked Sensor Systems},
pages = {159–171},
numpages = {13},
location = {Shenzhen, China},
series = {SenSys '18}
}

@inproceedings{10.1145/3210240.3210346,
author = {Li, Yan and Chi, Zicheng and Liu, Xin and Zhu, Ting},
title = {Chiron: Concurrent High Throughput Communication for IoT Devices},
year = {2018},
isbn = {9781450357203},
publisher = {Association for Computing Machinery},
address = {New York, NY, USA},
url = {https://doi-org.proxy.lib.ohio-state.edu/10.1145/3210240.3210346},
doi = {10.1145/3210240.3210346},
booktitle = {Proceedings of the 16th Annual International Conference on Mobile Systems, Applications, and Services},
pages = {204–216},
numpages = {13},
location = {Munich, Germany},
series = {MobiSys '18}
}

@INPROCEEDINGS{8486349,
  author={Wang, Wei and Xie, Tiantian and Liu, Xin and Zhu, Ting},
  booktitle={IEEE INFOCOM 2018 - IEEE Conference on Computer Communications}, 
  title={ECT: Exploiting Cross-Technology Concurrent Transmission for Reducing Packet Delivery Delay in IoT Networks}, 
  year={2018},
  volume={},
  number={},
  pages={369-377},
  doi={10.1109/INFOCOM.2018.8486349}}

@ARTICLE{9444204,
  author={Han, Dianqi and Li, Ang and Zhang, Lili and Zhang, Yan and Li, Jiawei and Li, Tao and Zhu, Ting and Zhang, Yanchao},
  journal={IEEE/ACM Transactions on Networking}, 
  title={Deep Learning-Guided Jamming for Cross-Technology Wireless Networks: Attack and Defense}, 
  year={2021},
  volume={29},
  number={5},
  pages={1922-1932},
  doi={10.1109/TNET.2021.3082839}}

@ARTICLE{10189210,
  author={Liu, Xin and Chi, Zicheng and Wang, Wei and Yao, Yao and Hao, Pei and Zhu, Ting},
  journal={IEEE/ACM Transactions on Networking}, 
  title={High-Granularity Modulation for OFDM Backscatter}, 
  year={2024},
  volume={32},
  number={1},
  pages={338-351},
  doi={10.1109/TNET.2023.3286880}}

@ARTICLE{10125074,
  author={Yao, Yao and Li, Yan and Zhu, Ting},
  journal={IEEE Transactions on Mobile Computing}, 
  title={Interference-Negligible Privacy-Preserved Shield for RF Sensing}, 
  year={2024},
  volume={23},
  number={5},
  pages={3576-3588},
  doi={10.1109/TMC.2023.3276930}}

@inproceedings {285483,
author = {Ang Li and Jiawei Li and Dianqi Han and Yan Zhang and Tao Li and Ting Zhu and Yanchao Zhang},
title = {{PhyAuth}: {Physical-Layer} Message Authentication for {ZigBee} Networks},
booktitle = {32nd USENIX Security Symposium (USENIX Security 23)},
year = {2023},
isbn = {978-1-939133-37-3},
address = {Anaheim, CA},
pages = {1--18},
url = {https://www.usenix.org/conference/usenixsecurity23/presentation/li-ang},
publisher = {USENIX Association},
month = aug
}

@inproceedings{10.1145/3395351.3399367,
author = {Chi, Zicheng and Li, Yan and Liu, Xin and Wang, Wei and Yao, Yao and Zhu, Ting and Zhang, Yanchao},
title = {Countering cross-technology jamming attack},
year = {2020},
isbn = {9781450380065},
publisher = {Association for Computing Machinery},
address = {New York, NY, USA},
url = {https://doi-org.proxy.lib.ohio-state.edu/10.1145/3395351.3399367},
doi = {10.1145/3395351.3399367},
booktitle = {Proceedings of the 13th ACM Conference on Security and Privacy in Wireless and Mobile Networks},
pages = {99–110},
numpages = {12},
location = {Linz, Austria},
series = {WiSec '20}
}

@inproceedings{10.1145/3769102.3770620,
author = {Song, Guanqun and Li, Yan and Zhu, Ting},
title = {A Metal Sensing and Biometric-based Tracking System},
year = {2025},
isbn = {9798400722387},
publisher = {Association for Computing Machinery},
address = {New York, NY, USA},
url = {https://doi-org.proxy.lib.ohio-state.edu/10.1145/3769102.3770620},
doi = {10.1145/3769102.3770620},
booktitle = {Proceedings of the Tenth ACM/IEEE Symposium on Edge Computing},
articleno = {23},
numpages = {17},
location = {the Hilton Arlington National Landing, Arlington, VA, USA},
series = {SEC '25}
}

\end{document}